\documentclass[preprint,12pt,3p]{elsarticle}

\usepackage{hyperref}
\hypersetup{
    colorlinks=true,
    linkcolor=blue,
    filecolor=magenta,      
}
    
\usepackage{times}
\usepackage{amssymb, amsmath, amsbsy}
\usepackage{makecell}  
\usepackage{graphicx}
\usepackage{float}
\usepackage{multirow}
\usepackage{xcolor}
\usepackage{soul}
\usepackage[capitalise]{cleveref}
\crefformat{appendix}{#2#1#3}

\usepackage{natbib}

\newcommand{\bs}[1]{\boldsymbol{#1}}
\newcommand{\mc}[1]{\mathcal{#1}}

\newcommand{\gradd}[0]{\bar{\bs{\nabla}}}

\newcommand{\parttd}[1]{\frac{\partial #1}{\partial \bar{t}}}

\newcommand{\pard}[2]{\frac{\partial #1}{\partial #2}}
\newcommand{\pardd}[2]{\frac{\partial^2 #1}{\partial #2 ^2}}

\usepackage{amsmath}

\DeclareMathOperator*{\argmin}{arg\,min}

\usepackage{algorithm}
\usepackage{algpseudocode}

\journal{arXiv}
\date{September 7, 2022}

\begin{document}

\begin{frontmatter}

\title{Inverse modeling of nonisothermal multiphase poromechanics using physics-informed neural networks}

\author[SHARIF]{\href{https://orcid.org/0000-0002-4352-5435}{\includegraphics[scale=0.06]{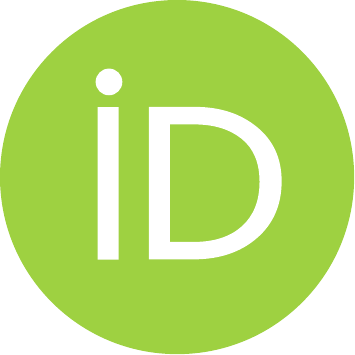}\hspace{1mm}Danial Amini}}
\address[SHARIF]{Sharif University of Technology}
\ead{aminidaniel.civil@gmail.com}

\author[MIT]{\href{https://orcid.org/0000-0003-2659-0507}{\includegraphics[scale=0.06]{orcid.pdf}\hspace{1mm}Ehsan Haghighat}}
\address[MIT]{Massachusetts Institute of Technology}
\ead{ehsanh@mit.edu}

\author[MIT]{\href{https://orcid.org/0000-0002-7370-2332}{\includegraphics[scale=0.06]{orcid.pdf}\hspace{1mm}Ruben Juanes}\corref{cor1}}
\ead{juanes@mit.edu}
\cortext[cor1]{Correspondence to: Ruben Juanes (juanes@mit.edu)}

\begin{abstract}
We propose a solution strategy for parameter identification in multiphase thermo-hydro-mechanical (THM) processes in porous media using physics-informed neural networks (PINNs). We employ a dimensionless form of the THM governing equations that is particularly well suited for the inverse problem, and we leverage the sequential multiphysics PINN solver we developed in previous work. We validate the proposed inverse-modeling approach on multiple benchmark problems, including Terzaghi's isothermal consolidation problem, Barry-Mercer's isothermal injection-production problem, and nonisothermal consolidation of an unsaturated soil layer. We report the excellent performance of the proposed sequential PINN-THM inverse solver, thus paving the way for the application of PINNs to inverse modeling of complex nonlinear multiphysics problems. 
\end{abstract}


\end{frontmatter}

\section{Introduction}\label{sec:intro}
Subsurface technologies like groundwater management \cite{sophocleous02, gorelickzheng15}, geothermal energy production \citep{miloratester76}, seasonal storage of natural gas \citep{jha2015reservoir} and hydrogen \citep{gabrielli20-uhs}, geologic carbon storage \citep{ipcc-co2, szulczewski12-pnas} and geological waste disposal \citep{veil2002drilling} rely on the injection and extraction of fluids underground. This, in turn, leads to evolving fluid pressures and temperature, as well as deformation and stress changes in the rock \citep{coussy2004poromechanics}. Such changes can lead to deleterious effects like surface subsidence \citep{gallowayburbey11}, seismicity \citep{nrc13-inducedseismicity, ellsworth13, keranen14} and fluid leakage across caprocks and along geologic faults \citep{bensegleeson13}. Thus, it is essential to develop quantitative models that enable understanding and prediction of these processes, to maximize the technology efficiency while minimizing hazards.

Mathematical models of these complex processes often take the form of coupled partial differential equations (PDEs), which describe (possibly nonisothermal and multiphase) flow through porous media and mechanics of geomaterials \citep{coussy2004poromechanics}. Over the past few decades, well-established computational methods have been developed for the numerical solution of these mathematical models \citep[e.g.,][]{lewis1998finite, zienkiewicz1999computational, armero1999formulation, settari1998coupled, thomas2003coupled, schrefler2004multiphase, jha2007locally, white2008stabilized, jha2014coupled}. Solving the inverse problem and identifying the parameters of the PDEs, however, continues to pose exceptional challenges due to the nonlinear nature of the equations, their strong coupling, and the extremely high dimensionality of the problem due to medium heterogeneity \citep{carrera05-inverseproblem, cuimarzouk14}. Given the ever-increasing availability of both static and dynamic data such as 3D seismic imaging, well logs, surface deformation from InSAR and GPS, well pressure and fluid compositions and downhole strain gauges, there is a growing interest in the direct integration of observation data in solution methods---and hence streamlined identification of PDE parameters, aiming to minimize the mismatch between sensor measurements and model predictions. Classical data assimilation and optimization techniques treat forward solvers as black-box estimators and therefore typically require an unfeasible number of forward simulations during the optimization. Physics-informed neural networks (PINNs), developed to extend the inherent capabilities of artificial neural networks to incorporate PDE constraints \citep{raissi2019physics}, offer the potential of a unified forward and inverse solver, and thus a framework to inherently combine measured data and model predictions. Here, we develop such a framework for forward and inverse modeling of THM problems using PINNs. 

The inverse problem in groundwater flow, transport, and reservoir geomechanics has a long history \citep[e.g.,][]{carreraneuman86-1, mclaughlintownley96, carrera05-inverseproblem, iglesias2012data}. Whether it is posed as a ``history matching'' problem \citep{lumley2001time, hesse2014joint, kang16-sequential, alghamdi2020bayesian} or a ``data assimilation'' problem \citep{vasco2001coupled, jha2015reservoir}, approaches are faced with fundamental challenges like ill-posedness, nonconvergence, and the ``curse of dimensionality'' \citep{oliver2011recent, iglesias2012data, cuimarzouk14}. The advent of machine learning (ML) tools that incorporate automatic differentiation \citep{baydin17}, such as TensorFlow \citep{abadi16-tensorflow} and PyTorch \citep{paszke19-pytorch}, has made it possible to perform optimization on deep and complex neural network architectures, and to simulate the outcomes of physical laws. In particular, \citet{raissi2019physics} introduced PINNs---a framework that leverages automatic differentiation to incorporate PDE constraints into ML optimization. An inherent advantage of PINN for the solution of inverse problems is that the forward model and the optimizer exchange information directly, using the value of the loss function and its gradient.

As discussed in \citet{karniadakis2021physics}, there are multiple bottlenecks associated with PINNs, including the training time of PINN forward solvers compared to classical methods, their suitability for large-scale problems with complex spatiotemporal domains, and their applicability to coupled nonlinear problems. The use of adaptive-weight strategies (e.g., \cite{chen2018gradnorm, wang2020understanding, wang2020ntk, wang2022respecting}) has significantly improved the training cost and the accuracy of multi-objective optimization, and the application of PINN solvers to complex domains has been addressed using time and space discretization methods using domain decomposition approaches \cite{5-PINN-cPINN, 33-PINN-XPINN}.

The PINN framework has been aggressively adopted and extensively explored for forward and inverse analysis and surrogate modeling of problems in fluid mechanics \cite{jin2021nsfnets,wu2018physics,cai2021physicsfluid,rao2020physics, mao2020physics}, solid mechanics \cite{haghighat2021physics, rao2021physics}, heat transfer \cite{cai2021physics, niaki2021physics}, and reactive flow problems  \cite{laubscher2021simulation} (for a detailed review see \cite{karniadakis2021physics,cuomo2022scientific}). It has also been applied recently in the area of subsurface flow and mechanics \cite{tartakovsky2020physics, fuks2020limitations, fraces2020physics, fraces2021physics, almajid2022prediction, shokouhi21-jch, kadeethum20-plosone, bekele2020physics, bekele2021physics, millevoi4074416physics}. For example, \citet{fuks2020limitations} and follow-up studies \citep{almajid2022prediction, fraces2020physics, fraces2021physics} tested PINN's robustness for inversion and forward modeling of two-phase fluid flow in a rigid matrix with various flux functions. \citet{bekele2021physics} identified the consolidation coefficient of Terzaghi's problem, but the predicted pressure distribution for the Barry-Mercer problem was not accurate \cite{bekele2020physics}. Based on single-phase fluid flow equation, \citet{harp2021feasibility} implemented a framework to manage the underground reservoir pressure by determining the fluid extraction rate. 

The application to coupled processes (such as hydromechanical or thermo-hydromechanical) has been nevertheless limited, in large part due to the challenges posed by network training. 
This bottleneck was recently addressed for poroelasticity problems \citep{haghighat2022physics} and thermo-poroelasticity problems \citep{amini2022physics} by means of a \emph{sequential} training strategy. These studies are the first to successfully and accurately solve the complex and nonlinear PDEs of (thermo-)poroelasticity using PINNs. 

Here, we build on these previous works to develop a solution strategy for parameter identification in multiphase thermo-hydro-mechanical (THM) processes. To this end, we cast the coupled governing equations in an appropriate dimensionless form that is well suited for inverse problems. We show that the stress-split sequential training \cite{haghighat2022physics} works effectively for the solution of the inverse problem. We apply the method to identify the model parameters in three benchmark problems of thermo-poromechanics: Terzaghi's consolidation problem, Barry-Mercer's injection--production problem, and consolidation in a stratum under nonisothermal conditions.

\section{Governing Equations}\label{sec:governings}
In this section, we present the governing equations of nonisothermal multiphase flow in poroelastic media, which include linear momentum balances, mass conservation for each fluid phase (assumed immiscible), and energy balance \cite{coussy2004poromechanics, amini2022physics}.
Since machine-learning frameworks improve their performance when working with normalized data, we present a dimensionless form of these equations, discussed in \cite{haghighat2022physics,amini2022physics}, and propose an adjustment that leads to a form that is more suitable for inverse problems. The dimensionless form of the equations effectively normalizes the unknown fields, i.e., displacement, fluid pressure, and temperature fields.  The formulation for single-phase fluid flow is not presented for brevity, but it could be derived by setting the saturation to unity and ignoring the nonwetting phase. 

\subsection{Porous Media}
Let us consider a porous matrix $\Omega \in \mathbb{R}^D$, with $D$ as the spatial dimension, with bulk volume $V$ and pore volume $V_v$. The porosity is the ratio of the volume of pores to the bulk volume, $\phi=V_v/V$. The connectivity of these pores plays an important role in the transport properties of the porous material. The connected pores of the matrix may be filled with wetting or non-wetting fluid phases. The amount of wetting or non-wetting fluids filling the pore space is described by their respective degrees of saturation, defined as $S_w = V_w/V_v$ and $S_n = V_n/V_v$. The density of the bulk is then evaluated as $\rho_b=(1-\phi)\rho_s+\phi(S_w\rho_w+S_n\rho_n)$, where $\rho_s, \rho_w, \rho_n$ are solid, wetting fluid, and non-wetting fluid densities, respectively. 

The porous matrix filled with a fluid mixture behaves mechanically as a composite material, and two extreme conditions can be considered. On one end, we may have drained conditions, where the pore fluid can move freely and an applied load on the matrix does not result in an increase in fluid pressure. In this case, the deformation of the matrix under the applied load is governed by the stiffness properties of the solid matrix, and in particular by the \emph{drained} bulk modulus of the solid matrix $K_{dr}$. On the other end, the fluid cannot move freely and the bulk response of the system is governed by the bulk stiffness of both the solid matrix and the pore fluid, and controlled by the \emph{undrained} bulk modulus $K_u = K_{dr} + \frac{K_f}{\phi}$ \cite{wang-tlp, coussy2004poromechanics}. 

\subsection{Mechanics of the Solid Skeleton}

Given a displacement field ${\boldsymbol{u}}$, under the assumption of small deformation, the matrix strain ${\boldsymbol{\varepsilon}}$ and its volumetric and deviatoric components, i.e., $\varepsilon_v$ and $\boldsymbol{e}$, are expressed as:
\begin{align}
{\boldsymbol{\varepsilon}} &= {({\boldsymbol{\nabla}}{\boldsymbol{u}}+{\boldsymbol{\nabla}}{\boldsymbol{u}}^{T})}/{2}, \\
{\varepsilon_v} &= {\text{tr}({\boldsymbol{\varepsilon}})}, \label{eq5}\\
{\boldsymbol{e}} &= {\boldsymbol{\varepsilon}} - \frac{\varepsilon_v}{3}{\mathbf{1}}. \label{eq6}
\end{align}
According to Biot's theory of thermo-poroelasticity, a change in the Cauchy stress tensor $\boldsymbol{\sigma}$ is expressed as a function of the strain tensor $\boldsymbol{\varepsilon}$ or equivalently its volumetric and deviatoric components, i.e, $\varepsilon_v$ and $\boldsymbol{e}$, a change in the equivalent pressure $p_E$, and a change in the temperature $T$ \citep{biot1941general}. This is expressed mathematically as  \cite{coussy2004poromechanics}
\begin{align}
{\delta\boldsymbol{\sigma}}= {K_{dr} \varepsilon_v \mathbf{1}} + {3}\left({\frac{1-2\nu}{1+\nu}}\right){K_{dr}}{\boldsymbol{e}} - b\delta p_E  \mathbf{1} - {\beta_s}{K_{dr}}{\delta T \mathbf{1}}, \label{eq9}
\end{align}
or equivalently:
\begin{align}
{\boldsymbol{\sigma}-\boldsymbol{\sigma_0}} = {K_{dr} \varepsilon_v \mathbf{1}} + {3}\left(\frac{1-2\nu}{1+\nu}\right){K_{dr}}{\boldsymbol{e}} - {b}(p_E - p_{E0}) {\mathbf{1}} - {\beta_s K_{dr} {(T - T_0)}}{\mathbf{1}}, \label{SolidConstitute}
\end{align}
where $K_{dr}$ is the drained bulk modulus and $\nu$ is the Poisson ratio. ${p_E}$ is the equivalent pore pressure that results from the multiphase flow in the pores (as discussed later). ${\beta_s}$ denotes the thermal expansion coefficient of the solid skeleton. The volumetric part of the stress tensor is expressed as 
\begin{align}
    {\delta \sigma_v} = {K_{dr}}{\varepsilon_v} - {b}\delta p_E - {\beta_s}{K_{dr}}{\delta T}.
\end{align}

The mechanical equilibrium of the system is governed by the linear momentum balance equation,
\begin{align}
{\boldsymbol{\nabla}}{\cdot}{\boldsymbol{\sigma}} + {\rho_b}{g}{\boldsymbol{d}} = {\mathbf{0}}, \label{SolidMain}
\end{align}
where ${g}$ represents the gravity acceleration in the direction ${\boldsymbol{d}}$. The balance of angular momentum implies the symmetry of the Cauchy stress tensor, i.e., ${\boldsymbol{\sigma^T} = \boldsymbol{\sigma}}$. Substituting the constitutive relation \eqref{SolidConstitute} into \eqref{SolidMain}, we can express the Navier relations of poroelasticity as
\begin{align}
\begin{split}
{K_{dr}}\boldsymbol{\nabla}{\varepsilon_v}+{\frac{1}{2}}\left({\frac{1-2\nu}{1+\nu}}\right){K_{dr}} {\boldsymbol{\nabla} {(\boldsymbol{\nabla}{\cdot}{\boldsymbol{u}})}} + {\frac{3}{2}}\left({\frac{1-2\nu}{1+\nu}}\right) {K_{dr}} {\boldsymbol{\nabla}{\cdot}{(\boldsymbol{\nabla u})}} \\
 - {b {\boldsymbol{\nabla}{p_E}}}
  - {\beta_s}{K_{dr}}{\boldsymbol{\nabla} {T}} + {\rho_b}{g}{\boldsymbol{d}} = {\boldsymbol{0}}.
\end{split} \label{SolidDimensional}
\end{align}

\subsection{Multiphase Flow in Porous Media}

Considering ${m_{\alpha}}$ as the mass of fluid phase $\alpha$ per unit bulk volume, ${\boldsymbol{w}_{\alpha}}$ as the fluid mass flux of phase $\alpha$, and ${f_{\alpha}}$ as a volumetric source term for phase $\alpha$, the mass conservation law for each phase $\alpha$ takes the form \cite{jha2014coupled}:
\begin{align}
{\frac{d m_{\alpha}}{d t}} + {\boldsymbol{\nabla}}{\cdot}{\boldsymbol{w}_{\alpha}} = {\rho_{\alpha}}{f_{\alpha}}, \label{FluidMain}
\end{align}
in which $m_{\alpha}$ and mass flux $\boldsymbol{w}_{\alpha}$ are expressed as  $m_{\alpha} = \rho_{\alpha} S_{\alpha} \phi(1+\varepsilon_v)$ and ${\boldsymbol{w}_{\alpha}} = {\rho_{\alpha}}{\boldsymbol{v}_{\alpha}}$, respectively, where ${\boldsymbol{v}_{\alpha}}$ is defined based on the multiphase extension of Darcy's law as:
\begin{align}
{\boldsymbol{v}_{\alpha}} = -\frac{k}{\mu_{\alpha}}{k_{\alpha}^r}{({\boldsymbol{\nabla}{p_{\alpha}}}-{\rho_{\alpha}}{g}{\boldsymbol{d}})}. \label{DarcyMain}
\end{align}
Here, ${k}$ denotes the intrinsic permeability of the medium, $k_{\alpha}^r$ and $\mu_{\alpha}$ are the relative permeability and viscosity of fluid phase $\alpha$, and ${p}_{\alpha}$ represents the pressure of fluid phase $\alpha$. 

As a constitutive relation, when a multiphase fluid occupies the pore space, the variation of fluid content of phase ${\alpha}$ is given as \cite{coussy2004poromechanics, jha2014coupled}:
\begin{align}
{\left(\frac{\delta m}{\rho}\right)_{\alpha}} = {b_{\alpha}}{\varepsilon_v} + \sum_{k}{N_{\alpha k}}{\delta p_{k}} - {\beta_{s,\alpha}}{\delta T}, \label{FluidConstitute}
\end{align}
in which 
\begin{align}
 {N_{nn}} &= -{\phi}{\frac{\partial S_w}{\partial p_c}} + {\phi}{S_n}{c_n} + {S_n^2}{N}, \\
 {N_{nw}} &= {N_{wn}} = {\phi}{\frac{\partial S_w}{\partial p_c}} + {S_n}{S_w}{N}, \\
 {N_{ww}} &= -{\phi}{\frac{\partial S_w}{\partial p_c}} + {\phi}{S_w}{c_w} + {S_w^2}{N}, \\
 N &= (b-\phi)/{K_s}, \\
 \beta_{s,\alpha} &= S_{\alpha}((b-\phi)\beta_s+\phi\beta_{\alpha}),  
\end{align} 
where $c_{\alpha}$ is the compressibility and $\beta_{\alpha}$ is the thermal expansion coefficient of fluid phase ${\alpha}$.  Finally, the mass conservation equation of fluid phases can be obtained, substituting \eqref{FluidConstitute} into \eqref{FluidMain}, as:
\begin{align}
{\sum_{k}{(N_{\alpha k}+\frac{b_{\alpha} b_k}{K_{dr}})}}{\frac{\partial p_k}{\partial t}} + {\frac{b S_{\alpha}}{K_{dr}}}{\frac{\partial \sigma_v}{\partial t}} - {\phi S_{\alpha}({\beta_{\alpha}} - {\beta_s})}{\frac{\partial T}{\partial t}}  + {\boldsymbol{\nabla}}{\cdot}{\boldsymbol{v}_{\alpha}} - {f_{\alpha}} = {0}, 
\label{FluidDimensional}
\end{align}

As discussed in \cite{coussy2004poromechanics}, two common approaches are used to account for the effect of multiphase fluid pressure on the deformation of the porous medium: the saturation-averaged pore pressure and the equivalent pore pressure. Only the latter leads to a thermodynamically consistent (entropy stable) formulation \cite{coussy2004poromechanics, kim2013rigorous}, so here we use the equivalent pore pressure, defined as:  
\begin{align}
{p_E} = {\sum_{\alpha}{S_{\alpha}}{p_{\alpha}}}-{U}, 
\end{align}
where $U$ is the interfacial energy, defined in incremental form as
\begin{align}
    {\delta U = \sum_{\alpha}{p_{\alpha}}{\delta {S_{\alpha}}}}.
\end{align}
In addition, when two fluid phases---wetting ($w$) and nonwetting ($n$)---occupy the pore space, capillary pressure ${p_c}$ is introduced as the pressure difference of nonwetting and wetting fluid pressures, as
\begin{align}
{p_c} = {p_n} - {p_w}.\label{pc_relation}
\end{align}
The capillary pressure is commonly assumed to be a function of the wetting phase saturation, $p_c=p_c(S_w)$.

\subsection{Heat Transfer in Porous Media}
As the last governing equation of THM processes in porous media, the energy balance equation is given as
\begin{align}
{\frac{d {m_\theta}}{d t}} + {\boldsymbol{\nabla}}{\cdot}{\boldsymbol{h}_{\theta}} = {G_\theta}, \label{HeatMain}
\end{align}
in which ${m_\theta}$ is the energy per unit bulk volume, ${\boldsymbol{h}_{\theta}}$ represents the heat flux, and ${G_\theta}$ denotes the volumetric heat source. In addition, we have the following constitutive laws for the energy balance equation:
\begin{align}
& {m_{\theta}} = {(\rho C)_\text{avg}}{T}, \label{HeatConstitute1}\\
& {\boldsymbol{h}_{\theta}} = ({\rho_n}{C_n}{\boldsymbol{v}_n} + {\rho_w}{C_w}{\boldsymbol{v}_w}){T} -{\lambda_\text{avg}}{\boldsymbol{\nabla}}{T},\label{HeatConstitute2}
\end{align}
with ${(\rho C)_\text{avg}}$ as the average heat capacity and ${\lambda_\text{avg}}$ as the average thermal conductivity of the porous medium, given as:
\begin{align}
& (\rho C)_\text{avg}={(1-\phi)\rho_s C_s } + \sum_{\alpha}{\phi S_{\alpha} \rho_{\alpha} C_{\alpha}}, \\
& {\lambda_\text{avg}} = {(1-\phi)}{\lambda_s} + \sum_{\alpha}{\phi}{S_{\alpha}}{\lambda_{\alpha}}. \label{eq22}
\end{align}
Here, $C_s, \lambda_s$ and $C_{\alpha}, \lambda_{\alpha}$ express the heat capacity and thermal conductivity of solid $s$ and fluid phase ${\alpha}$. 
The equation that describes the heat transfer in multiphase poroelasticity is finally expressed, by replacing \eqref{HeatConstitute1} and \eqref{HeatConstitute2} into \eqref{HeatMain}, as:
\begin{align}
{(\rho C)_\text{avg}}\frac{\partial T}{\partial t} +{{(\rho_n C_n \boldsymbol{v}_n + \rho_w C_w \boldsymbol{v}_w) {\cdot}{\boldsymbol{\nabla}}{T}} - {\boldsymbol{\nabla}}{\cdot}({\lambda_\text{avg}}{\boldsymbol{\nabla}{T}}}) - {G_\theta} = {0}. \label{HeatDimensional}
\end{align}

\subsection{Dimensionless Governing Relations} 
As discussed, machine learning tools are optimized to work with normalized data. Therefore, to facilitate the optimizer's task of training a PINN solver, we normalize the unknown field variables and parameters, as discussed at length in our previous studies on forward PINN solutions of multiphysics problems \cite{haghighat2022physics, amini2022physics}. Since we are now focused on inverse PINN solvers, we found that a modification was needed to have normalized trainable parameters as PDE coefficients. Below, we propose a slight modification to the previously discussed relations that are more suitable for inverse problems. 

We define the dimensionless variables (noted with an overbar) as:
\begin{align}
& {\bar{t}} = \frac{t}{t^*},~~~{\bar{x}} = \frac{x}{x^*},~~~{\boldsymbol{\bar{u}}} = \frac{\boldsymbol{u}}{u^*}, ~~~{\bar{\boldsymbol{\varepsilon}}} = \frac{\boldsymbol{\varepsilon}}{{\varepsilon}^*},~~~{\bar{p}}_{\alpha} = \frac{p_{\alpha}}{p^*},~~~{\bar{\boldsymbol{\sigma}}} = \frac{\boldsymbol{\sigma}}{p^*},~~~{\bar{T}} = \frac{T}{T^*},
\end{align}
where, $t^*, x^*, u^*, \varepsilon^*, p^*$, and $T^*$ are characteristic (normalizing) factors for time, spatial dimension, displacement, strain, pore pressure and temperature, respectively. These normalizing variables are selected such that the problem is mapped to a domain close to unity. For instance, $t^*$ is selected as the final solution time; therefore dimensionless time $\bar{t}$ is mapped onto $[0, 1]$. Additionally, let us introduce the new normalized THM parameters:
\begin{equation}
\begin{split}
& {\bar{M}} = \frac{M}{{K^*_{dr}}},~~~{\bar{K}_{dr}} = \frac{K_{dr}}{{K^*_{dr}}},~~~{\bar{k}} = \frac{k}{k^*},~~~{\bar{\mu}_{\alpha}} = \frac{\mu_{\alpha}}{\mu^*}, \\
& {\bar{\rho}_{\alpha}} = \frac{\rho_{\alpha}}{\rho^*},~~~{\bar{\lambda}} = \frac{\lambda}{\lambda^*},~~~ {\bar{C}_{\alpha}} = \frac{C_{\alpha}}{{C^*}},~~~ {\bar{N}_{\alpha k}} = {N_{\alpha k}}{K^*_{dr}},
\end{split} \label{eq27}
\end{equation}
where, $K^*_{dr}, k^*, \mu^*, \rho^*, \lambda^*, \text{and}~C^*$ are some average and normalizing scalars for drained bulk modulus, intrinsic permeability, fluid viscosity, fluid density, thermal conductivity and heat capacity, respectively. 

Note that in our previous work \cite{haghighat2022physics, amini2022physics}, we derived these parameters by factorizing PDE coefficients after replacing the dimensionless variables \eqref{eq27} in the THM PDEs. Here, we normalize PDE parameters \emph{a priori} so that their dimensionless (overbar) coefficients remain in the final PDE for inversion. 

The dimensionless form of the linear momentum equation and stress--strain relation given in \eqref{SolidDimensional} are given as:
\begin{align}
\begin{split}
    & {\bar{K}_{dr}}\boldsymbol{\bar{\nabla}}{\bar{\varepsilon}_v}+{\frac{1}{2}}\left({\frac{1-2\nu}{1+\nu}}\right){\bar{K}_{dr}}{\boldsymbol{\bar{\nabla}}}{(\boldsymbol{\bar{\nabla}}{\cdot}{\boldsymbol{\bar{u}}})} + {\frac{3}{2}}\left({\frac{1-2\nu}{1+\nu}}\right){\bar{K}_{dr}} {\boldsymbol{\bar{\nabla}}}{\cdot}{(\boldsymbol{\bar{\nabla}}{\boldsymbol{\bar{u}}})} - {b}{\sum_{\alpha}}{\boldsymbol{\bar{\nabla}}}{({S_{\alpha}}{\bar{p}_{\alpha}})} \\
&~~~~~~~~~~~~~~~~~~ - {\bar{\beta}_s}{\bar{K}_{dr}}{N_T}{\boldsymbol{\bar{\nabla}}}{\bar{T}} + {\bar{\rho}_b}{N_d}{\boldsymbol{d}} = \boldsymbol{0},
\end{split} \label{eq28} \\
\begin{split}
& {\boldsymbol{\bar{\sigma}}-\boldsymbol{{\bar{\sigma}}_0}} = {\bar{K}_{dr}}{\bar{\varepsilon}_v} {\mathbf{1}} + 3\left({\frac{1-2\nu}{1+\nu}}\right){\bar{K}_{dr}}{\boldsymbol{\bar{e}}} - {b}\sum_{\alpha}{S_{\alpha} ({\bar{p}_{\alpha}} - {{\bar{p}_{\alpha 0}}})} {\mathbf{1}} \\
&~~~~~~~~~~~~~~~~~~ - {\bar{\beta}_s}{\bar{K}_{dr}}{{N_T} {(\bar{T} - {\bar{T}}_0)}}{\mathbf{1}}, 
\end{split} \label{eq29} \\
& {\delta {\bar{\sigma}}_v} = {\bar{K}_{dr}}{\delta {\bar{\varepsilon}}_v} -{b}{\sum_{\alpha}}{S_{\alpha}}{\delta{\bar{p}_{\alpha}}} - {\bar{\beta}_s}{\bar{K}_{dr}}{N_T}{\delta {\bar{T}}}. \label{eq30}  
\end{align}

\noindent
The mass conservation for fluid phases, relation \eqref{FluidDimensional}, and Darcy's law, relation \eqref{DarcyMain}, are expressed in dimensionless form as:
\begin{align}
\begin{split}
& {\sum_{k} ({\bar{N}_{\alpha k}}+\frac{b_{\alpha} b_k}{\bar{K}_{dr}}){D^*}{\frac{\partial \bar{p}_k}{\partial \bar{t}}}} + {\frac{b S_{\alpha}}{\bar{K}_{dr}}}{D^*} \frac{\partial \bar{\sigma}_v}{\partial \bar{t}} - {\phi S_{\alpha}}\left({\bar{\beta}_{\alpha}}-{\bar{\beta}_s}\right){Q^*} \frac{\partial \bar{T}}{\partial \bar{t}} - {\boldsymbol{\bar{\nabla}}}\cdot{\bar{\bs{v}}_{\alpha}} - {f_{\alpha}^*} = {{0}}, 
\end{split} \\
& {\bar{\bs{v}}_{\alpha}} = {\frac{\bar{k}}{\bar{\mu}_{\alpha}}} {k_{\alpha}^r}{({\boldsymbol{\bar{\nabla}} {\bar{p}_{\alpha}}}-{\bar{\rho}_{\alpha}}{N_d}{\boldsymbol{d}})}.
\end{align}

\noindent
Finally, the energy balance equation \eqref{HeatDimensional} can be rewritten in the non-dimensional format as:
\begin{align}
{H^*}{\frac{\partial \bar{T}}{\partial \bar{t}}} + {J^*}{\left({\bar{\rho}_n}{\bar{C}_n}{\bar{\bs{v}}_n}
+ {\bar{\rho}_w}{\bar{C}_w}{\bar{\bs{v}}_w} \right)} {\cdot}{\boldsymbol{\bar{\nabla}}}{\bar{T}}- {F^*}{\boldsymbol{\bar{\nabla}}}{\cdot}{({{\bar{\lambda}_\text{avg}}}{\boldsymbol{\bar{\nabla}}}{\bar{T}})} - {G_\theta^*} = {0}, \label{eq32}
\end{align}
Lastly, the dimensionless parameters are
\begin{align}
\begin{split}
& {u^*} = {\frac{p^*}{K^*_{dr}}}{x^*},~ {\varepsilon^*} = {\frac{u^*}{x^*}}, ~{N_T} = {\beta^*}{K^*_{dr}}\frac{T^*}{p^*},~{N_d} = {\frac{x^* \rho^*}{p^*}}{g}, \\
&{D^*}=\frac{{\mu^*}{x^*}^2}{K^*_{dr} {k^*}{t^*}},~{Q^*} = \frac{{\beta^*}T^* {\mu^*}{x^*}^2}{{t^*}{k^*}{p^*}}, ~{f_{\alpha}^*} = {f_{\alpha}}\frac{\mu^* {x^*}^2}{k^* p^*},\\
&{H^*} = {\frac{(\rho C)_\text{avg}}{\rho^* C^*}},~ {J^*} = {\frac{k^* {p^*}{t^*}}{\mu^* {x^*}^2}}, ~ {F^*} = \frac{{t^*}{\lambda^*}}{{\rho^* C^*}{x^*}^2},~ {G_\theta^*} = {G_\theta}{\frac{{t^*}}{T^* {\rho^* C^*}}}.
\end{split}
\end{align}


\section{PINN-THM Forward and Inverse Solver}

The physics-informed neural network framework used here to solve inverse poroelasticity problems is based on the sequential solution strategy proposed in \citet{haghighat2022physics}. In that study, we explored both simultaneous and sequential solvers, and we considered different neural network architectures as well as adaptive-weight schemes. 
Since most of the details remain almost the same, we skip re-writing the details and point the interested reader to the studies we reported in \cite{haghighat2022physics,amini2022physics} for solving HM and THM processes in porous media, and therefore, we only present the essential details for the completeness of our discussions. 

\begin{figure}[t]
\centering
\includegraphics[width=1\textwidth]{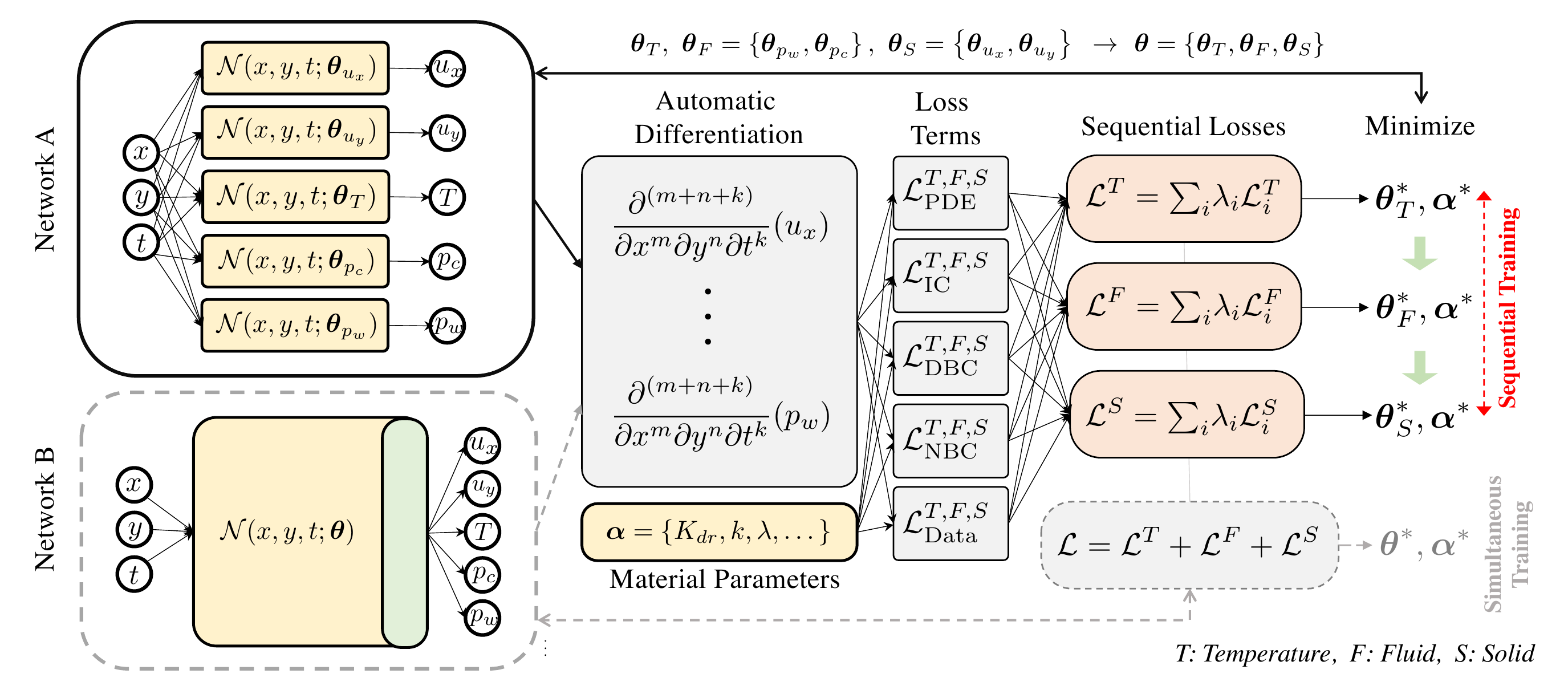}
\caption{PINN--THM network architecture schematic. Two network types can be used: (A) separate networks for each solution variable (no parameter sharing), and (B) a single network with multiple outputs (shared parameters). The yellow colored boxes present the network parameters. Differential operators are evaluated using the automatic differentiation algorithm. Characterization parameters are listed in the trainable weights $\boldsymbol{\alpha}$. The loss residuals associated with the temperature (T), fluid (F), and solid (S) governing differential equations ($\mc{L}^{T,F,S}_{\text{PDE}}$), Dirichlet ($\mc{L}^{T,F,S}_{\text{DBC}}$) and Neumann ($\mc{L}^{T,F,S}_{\text{NBC}}$) boundary conditions, initial condition ($\mc{L}^{T,F,S}_{\text{IC}}$), and data ($\mc{L}^{T,F,S}_{\text{Data}}$) are evaluated on sampling points ($\textbf{X},\textbf{Y},\textbf{T}$). The temperature, fluid, and solid total losses $\mathcal{L}^{T,F,S}$ are finally evaluated as the weighted summation of individual loss terms. Minimizing the total loss results in the optimal network parameters. Sequential and simultaneous training strategies can be used. However, using network architecture A becomes mandatory for the sequential training strategy, in order to freeze one network during sequential iterations.}
\label{figs:pinn_architecture}
\end{figure}

The PINN--THM network architecture is shown schematically in \cref{figs:pinn_architecture}. Two network architectures can be used in general. Separate networks for each solution variable (Network A) or a single network with multiple outputs (Network B). We have explored these architectures in the past and found Network~A to be a better strategy (see \cite{haghighat2021physics, haghighat2022physics} for details). Additionally, for a sequential training strategy, Network A becomes the only choice because one can freeze any set of parameters during sequential iterations. The trainable parameters of the PDEs (those that we are interested in characterizing given data) are listed in $\bs{\alpha}$. The automatic differentiation algorithm is used to evaluate differential operators on each solution variable. The loss terms are then evaluated on the collocation points $\textbf{X},\textbf{Y},\textbf{T}$. Total loss is then evaluated as the weighted summation of the individual loss terms. We explored, in details, different adaptive weight schemes for the PINN--THM solver \cite{haghighat2022physics} and found that GradNorm \cite{chen2018gradnorm} results in the best accuracy and therefore is used in this study. The optimization is then performed using the ADAM optimizer. 

We leverage the stress-split sequential optimization strategy proposed in \cite{haghighat2022physics,amini2022physics}. To this end, we first optimize for temperature field by freezing the displacement and pressure networks. We then use the temperature results to optimize for the pressure network. We finally optimize for displacements given temperature and fluid pressure distributions using the fixed-stress operator split formulation \cite{kim2011stability, kim2013rigorous}. We repeat this sequence until convergence. The algorithm searches for optimal network parameters as well as PDE parameters simultaneously. Therefore, the forward and inverse solvers remain almost identical---a clear advantage of PINN solvers. Accordingly, the sequential optimization problem can be summarized as:
\begin{align}
&\bs{\theta}_{T}^{i+1}, \bs{\alpha}^{i+1} = \argmin_{\bs{\theta}_{T},\bs{\alpha}} \mathcal{L}_T (\mathbf{X}, \mathbf{Y}, \mathbf{T}, \bar{T}; \hat{P}_w^{i}, \hat{P}_c^i, \hat{U}_x^i, \hat{U}_y^i, \bs{\alpha}^{i}), \\
&\bs{\theta}_{F}^{i+1}, \bs{\alpha}^{i+1} = \argmin_{\bs{\theta}_{p_w,p_c},\bs{\alpha}} \mathcal{L}_F (\mathbf{X}, \mathbf{Y}, \mathbf{T}, \bar{P}_{w}; \hat{T}^{i+1}, \hat{U}_x^i, \hat{U}_y^i, \bs{\alpha}^{i+1}), \\
&\bs{\theta}_{S}^{i+1}, \bs{\alpha}^{i+1} = \argmin_{\bs{\theta}_{S},\bs{\alpha}} \mathcal{L}_S(\mathbf{X}, \mathbf{Y}, \mathbf{T}, \bar{U}_{x}, \bar{U}_y; \hat{T}^{i+1}, \hat{P}_w^{i+1}, \hat{P}_c^{i+1},\bs{\alpha}^{i+1}),
\end{align}
where, subscripts $T, F, S$ stand for thermal, fluid, and solid terms of the loss and network parameters, and $i$ represents the sequential iteration loop. The overbar terms $\bar{T},\bar{P_w},\bar{U}_x,\bar{U_y}$ highlight any given data for the inverse solver. The vectors $\hat{T}, \hat{P}_w, \hat{P}_c, \hat{U_x}, \hat{U}_y$ represent the approximate network outputs at collocation points $\mathbf{X},\mathbf{Y},\mathbf{T}$, getting updated during sequential iterations. Note that compared to PINN forward solvers, PDE parameters, listed in $\boldsymbol{\alpha}$, are also trainable for an inverse solver and are optimized during the optimization. Here, we form the loss functions using the dimensionless relations~\eqref{eq27}--\eqref{eq32}.

Even though some problems, such as Terzaghi's consolidation problem, are one-way coupled in their forward setup, they still require sequential iterations in their inverse analysis since their characteristic parameters are altered by all coupled processes.

\section{Applications}
In this section, we evaluate the performance of the proposed PINN inverse solver for the consolidation problem under (i)~isothermal and (ii)~nonisothermal conditions, and (iii)~the Barry-Mercer's injection-production problem. The parameters for characterization include permeability ${\bar{k}}$, drained bulk modulus ${\bar{K}_{dr}}$ and thermal conductivity ${\bar{\lambda}}$. We assume that Poison's ratio is known \emph{a priori}; therefore, one can evaluate Young's modulus using isotropic elasticity relations. Since these are normalized variables, we select their initial values as 2, with their expected target values from optimization as 1. The dimensional form of these variables are, however, important ultimately and are calculated using relations~\eqref{eq27}.

With regard to the network and optimization hyper-parameters and implementation details, we use the ADAM optimizer of TensorFlow/Keras package using the SciANN API \cite{haghighat2021sciann}. The batch size is set to 500 and each sequential training is performed using 5,000 epochs. 


\subsection{Terzaghi's Consolidation Problem}

\begin{figure}[H]
    \centering
    \includegraphics[width=0.4\textwidth]{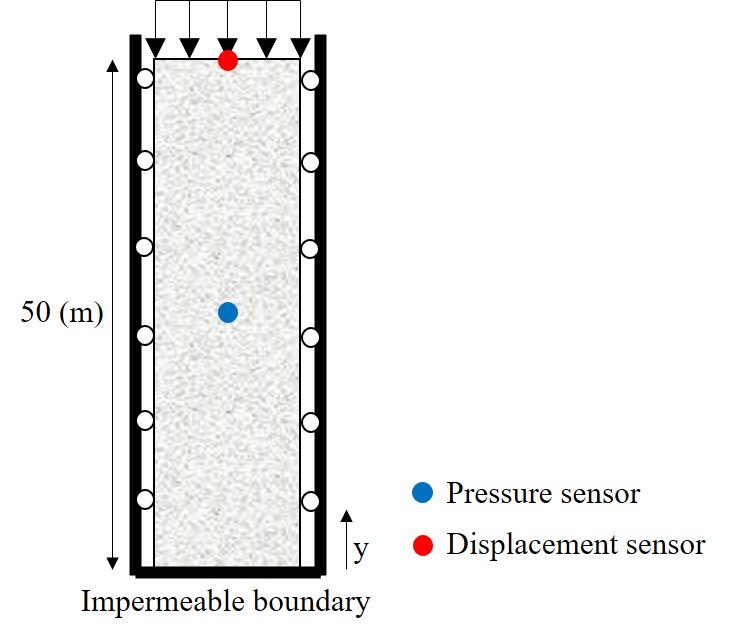}\caption{Terzaghi's consolidation problem; geometry, boundary conditions, and location of pressure (blue) and displacement (red) sensors.}
    \label{figs:Terzaghi-geometry}
\end{figure}

As the first example, let us consider the one-dimensional Terzaghi's consolidation problem. Consider a soil column of length $L=50$~m, constrained laterally and at the bottom, with no-flux condition on all sides except for the top face. The sample has a Young modulus $120~\text{MPa}$, drained Poisson's ratio $0.25$, Biot coefficient $1.0$, and permeability of $10^{-12}~\mathrm{m}^2$. The sample is suddenly subjected to an overburden stress $q=0.1~\text{MPa}$. The Biot's bulk modulus, ${M}$, expressed as $M^{-1} = \phi_0 c_f + ({b-\phi_0})/{K_s}$, is taken as $3 \times 10^5 ~\text{MPa}$. The one-dimensional single phase governing relations are therefore written as:
\begin{align}
& \left(\frac{b^2}{\bar{K}_{dr}} + \frac{1}{\bar{M}}\right){D^*}\parttd{\bar{p}} - {\frac{\bar{k}}{\bar{\mu}}} \frac{\partial^ 2\bar{p}}{\partial \bar{y}^2} = 0, \label{eqs:Terzaghi_fluid}\\ 
& {\bar{K}_{dr}}\pard{\bar{\varepsilon}_v}{\bar{y}} +{\frac{1}{2}}\left({\frac{1-2\nu}{1+\nu}}\right){\bar{K}_{dr}}\pardd{\bar{u}}{\bar{y}} + {\frac{3}{2}}\left({\frac{1-2\nu}{1+\nu}}\right){\bar{K}_{dr}} \pardd{\bar{u}}{\bar{y}} - {b}\pard{\bar{p}}{\bar{y}} = 0.
 \label{eqs:nondim1}
\end{align}
Here, we consider that the pressure and displacement data are given at two locations, one in the middle of the column for pressure, and another at the top for displacement, as shown in \cref{figs:Terzaghi-geometry}. The displacement and pressure networks are 100 neurons wide, each with 4 hidden layers and with hyperbolic-tangent activation function. A structured sampling grid of 41$\times$101 is used to train the problem. 

\begin{figure}[H]
    \centering
    \includegraphics[width=1\textwidth]{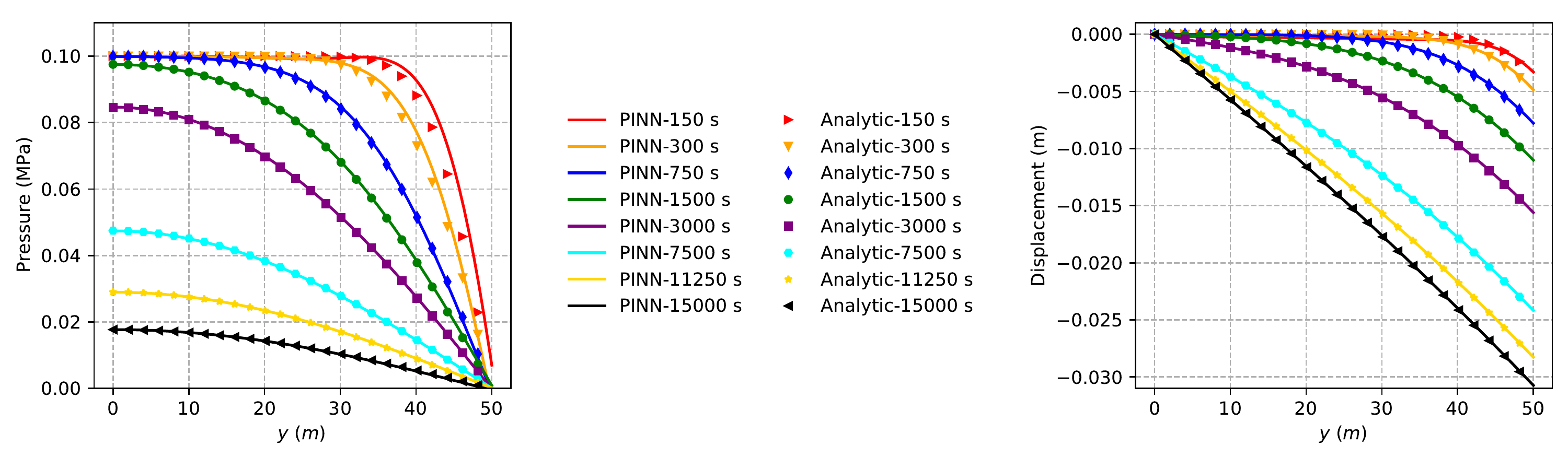}\caption{Terzaghi's consolidation problem; (Left) PINN's pressure solution, and (Right) PINN's displacement solution at different time steps, compared against the analytical solution.}
    \label{figs:Terzaghi_results}
\end{figure}

\begin{figure}[H]
    \centering
    \includegraphics[width=0.65\textwidth]{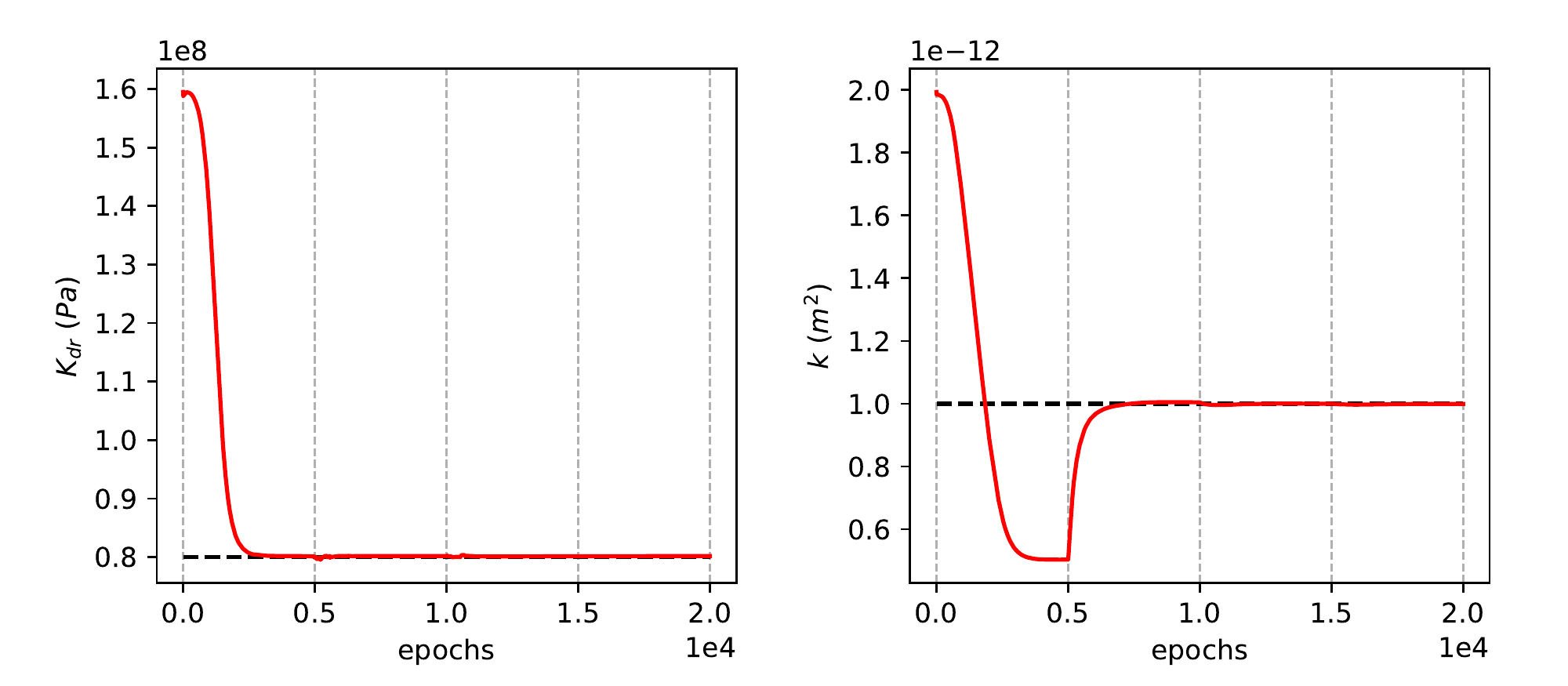}\caption{Terzaghi's consolidation problem; The characterization results (inverse solution) for bulk compressibility ${K_{dr}}$ and permeability ${k}$, as a function of training epochs, with vertical lines highlighting the sequential iterations.}
    \label{figs:Terzaghi_parameter}
\end{figure}

The results for displacement and pressure distributions are shown in \cref{figs:Terzaghi_results}, and the parameter identification results are plotted in \cref{figs:Terzaghi_parameter} with each vertical line highlighting the sequential iterations. 
Following the forward PINN--Poroelasticity solver \cite{haghighat2022physics}, we first invert for fluid flow, then use the output pressure to invert for solid deformation.  Since parameters are unknown \emph{a priori}, every sequential iteration may alter both solution variables and PDE parameters until convergence is reached.

As illustrated in \cref{figs:Terzaghi_parameter}, the permeability is first under-estimated due to the lack of information for the drained bulk modulus. With more accurate information on the drained bulk modulus, as a result of solving the solid PDE given the pressure distribution, this error is corrected during the following sequential iterations. The results show the remarkable performance of the sequential PINN solver for parameter identification in Terzaghi's consolidation problem.  



\subsection{Barry--Mercer's Injection--Production Problem}
As the second example, we characterize the drained bulk modulus and the permeability of a rectangular domain, subjected to a time-dependent point fluid injection--production, known as Barry--Mercer problem \cite{Barry-Mercer}. We use the setup of \cite{phillips2007coupling1,phillips2007coupling2}, with the injection/production well located at ${(x_0,y_0) = (0.25,0.25)}$, inside a two-dimensional domain of length and width as  ${a=b=1\text{m}}$. The solid and fluid phases are incompressible, the Biot coefficient is ${b=1}$, the elastic modulus and Poisson ratio are ${E=4.67~\text{MPa}}$ and ${\nu=0.167}$, respectively, and the initial pressure and displacement are set to zero. 

The injection/production function $f(t)$ is given as:
\begin{align}
f(t) = {2 \beta}{\delta(x-x_0)\delta(y-y_0)}{\sin(\beta t)},
\end{align}
with $\beta=0.5~\text{s}^{-1}$, that is associated with ${k=10^{-10}~\text{m}^2}$ and ${\mu=10^{-3}~\text{Pa.s}}$, given ${\beta}$'s definition \cite{phillips2007coupling1,phillips2007coupling2,haghighat2022physics}.
The Dirac delta function ${\delta}$ is approximated as ${\delta(x)}=\frac{1}{\alpha \sqrt{\pi}}{e^{-(x/\alpha)^2}}$, where we use ${\alpha=0.04}$. 
Similar to \citet{Barry-Mercer}, we normalize the temporal domain by defining ${\hat{t} = \beta t}$, where ${\hat{t} \in [0,2\pi]}$ is the dimensionless time. The single-phase two-way coupled HM governing equations are then summarized as:
\begin{align}
& \frac{b^2}{\bar{K}_{dr}}{D^*}\parttd{\bar{p}} + \frac{b}{\bar{K}_{dr}}{D^*}\parttd{\bar{\sigma}_v} - {\frac{\bar{k}}{\bar{\mu}}}\gradd^2 \bar{p} - {f^*} = 0, \label{eqs:Barry_fluid}\\
& {\bar{K}_{dr}}\boldsymbol{\bar{\nabla}}{\bar{\varepsilon}_v}+{\frac{1}{2}}\left({\frac{1-2\nu}{1+\nu}}\right){\bar{K}_{dr}}{\boldsymbol{\bar{\nabla}}}{(\boldsymbol{\bar{\nabla}}{\cdot}{\boldsymbol{\bar{u}}})} + {\frac{3}{2}}\left({\frac{1-2\nu}{1+\nu}}\right){\bar{K}_{dr}} {\boldsymbol{\bar{\nabla}}}{\cdot}{(\boldsymbol{\bar{\nabla}}{\boldsymbol{\bar{u}}})} - {b}{\boldsymbol{\bar{\nabla}}}{{\bar{p}}} = \boldsymbol{0}.
 \label{eqs:nondim2}
\end{align}

\begin{figure}[H]
    \centering
    \includegraphics[width=0.9\textwidth]{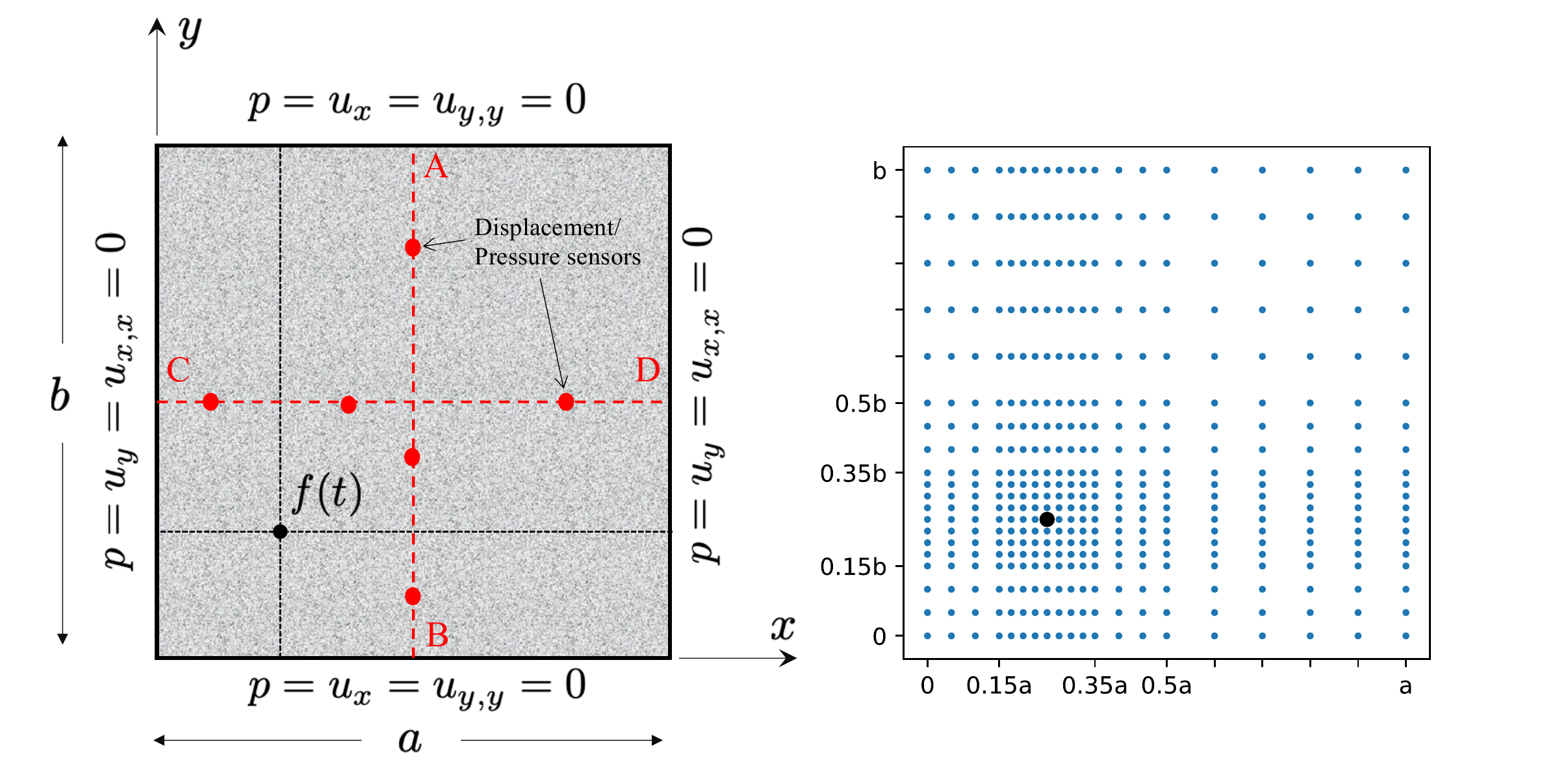}
    \caption{Barry--Mercer's injection--production problem. All edges are subjected to zero-pressure condition, and they are fixed in their normal direction. A production/injection source term $f(t)$ is applied at $(x_0, y_0)$. Red points, located at $0.1,~0.4,~\text{and}~0.8$ relative to a and b, along AB and CD lines represent the location of the sensors providing pressure and displacement data. The collocation grid is shown in blue in the right figure. Due to the type of the injection/production delta function, we use a finer grid around the well location. }
    \label{figs:barry_mercer_geometry}
\end{figure}

\begin{figure}[H]
    \includegraphics[width=1\textwidth]{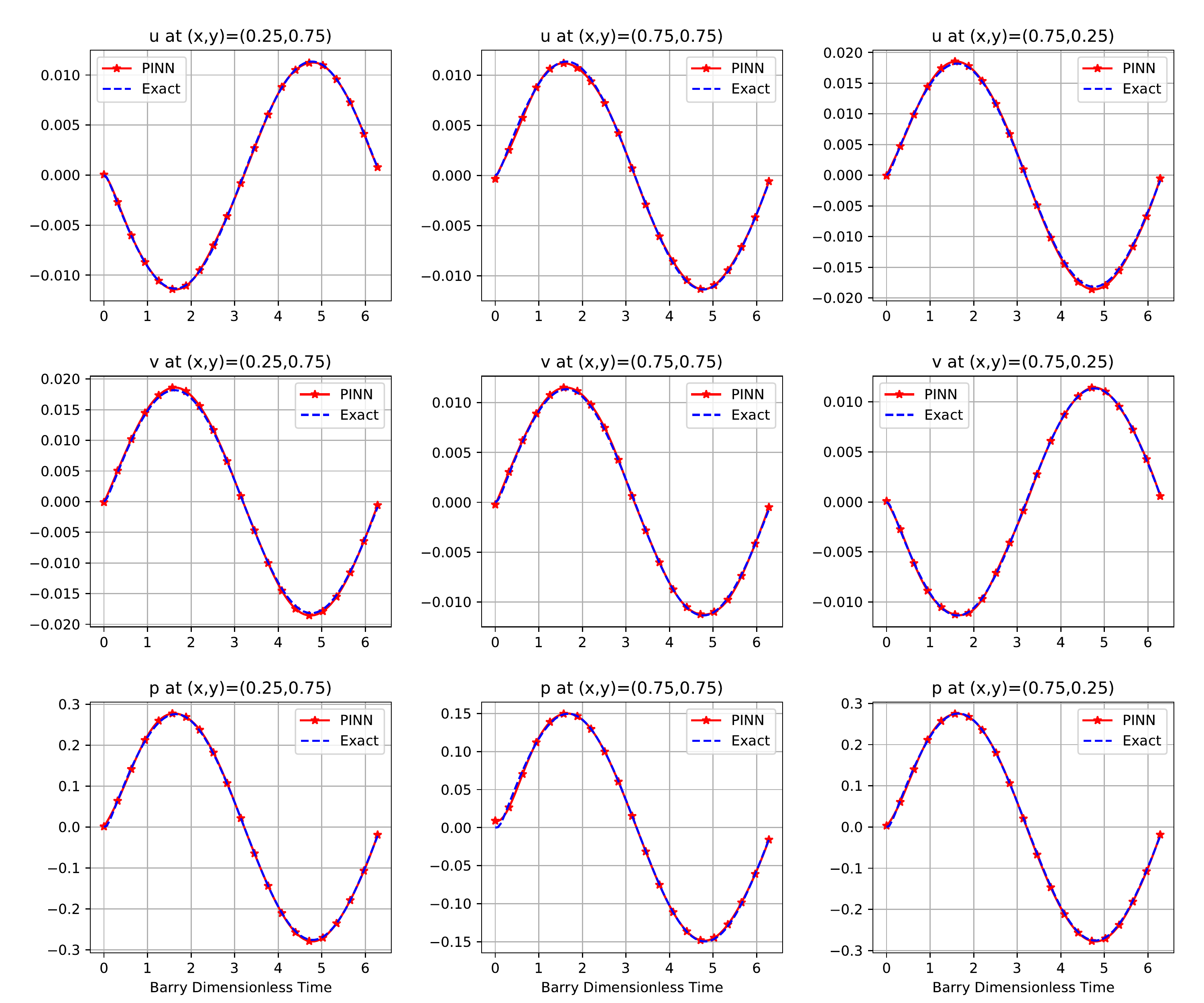}
    \caption{Barry--Mercer's PINN solution; (Top) horizontal displacement; (Middle) vertical displacement; (Bottom) pore pressure. Each column represent a different location away from the injection--production well.}
    \label{figs:barry_mercer_linePlot}
\end{figure}

\begin{figure}[H]
    \centering
    \includegraphics[width=0.65\textwidth]{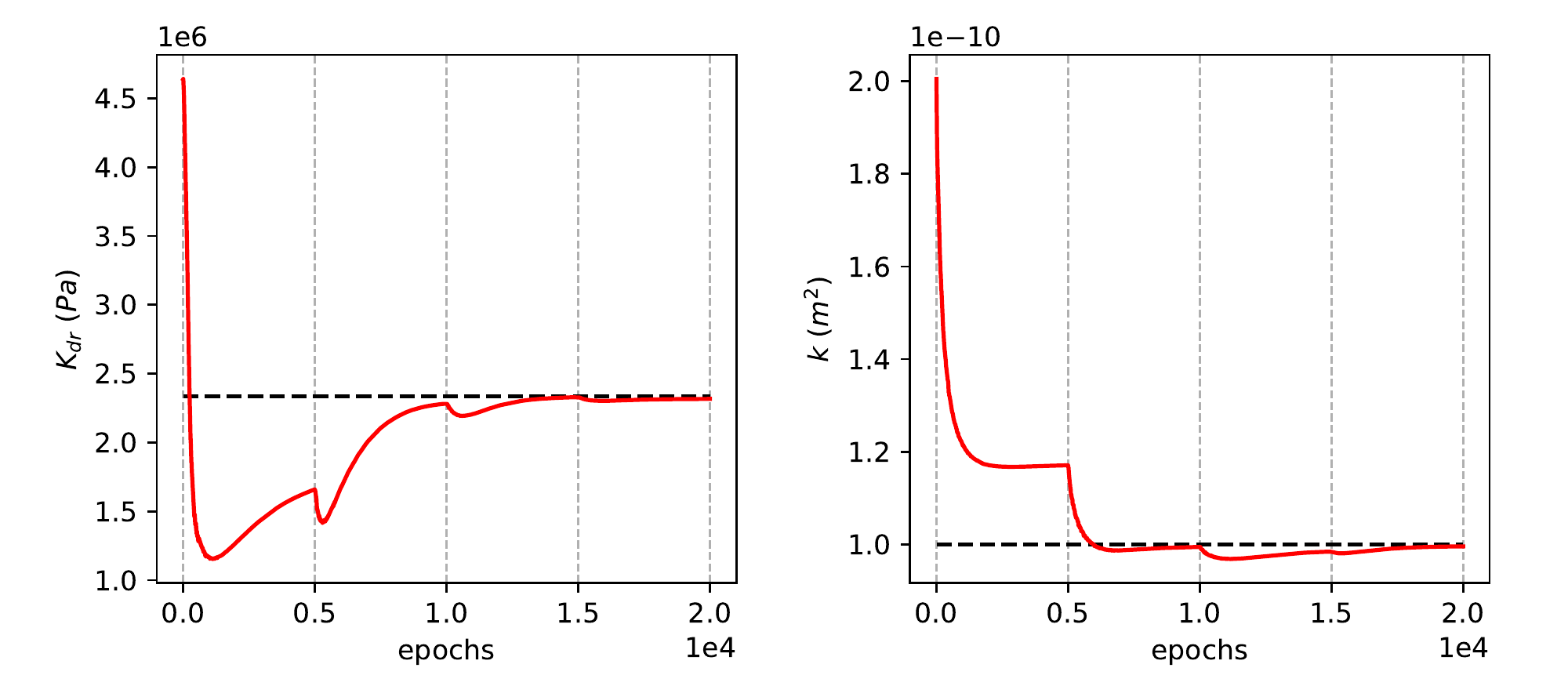}
    \caption{Barry--Mercer's solution; evolution of characterization of bulk compressibility ${K_{dr}}$ and permeability ${k}$ as a function of training epochs, with vertical lines highlighting the sequential iterations.}
    \label{figs:barry_mercer_parameter}
\end{figure}

Although not restrictive, we assume that vertical and horizontal displacements, as well as pore pressure, are given at a total of 6 sensors located along AB and CD lines shown in \cref{figs:barry_mercer_geometry}.
Additionally, since the injection and production source term is calculated based on a sharp Gaussian approximation of the Dirac delta function, with its spike localized near the injection well, we found that using more collocation points near the injection well increases the accuracy of the PINN solver. 
In \cref{figs:barry_mercer_linePlot} we present the obtained pressure and displacements at $(0.25, 0.75)m$, $(0.75, 0.75)m$, and $(0.75, 0.25)~m$, all away from the AB and CD lines where sensor data are provided, and the results match the expected analytical curves remarkably well. The evolution of the inversion parameters are plotted in \cref{figs:barry_mercer_parameter}, with vertical lines highlighting the epoch of sequential iterations. Again, the first sequential iteration has the highest error, with each sequential iteration resulting in improved accuracy, until reaching very high accuracy at the fifth iteration of sequential training. The displacement and pressure networks are 100 neurons wide, each with 4 hidden layers and with hyperbolic-tangent activation function and with Fourier features. A structured sampling grid of 21$\times$21$\times$41 is used to train the problem. 

\subsection{Thermoelastic Consolidation of Unsaturated Stratum}
As the last example, we consider a fully coupled THM problem under nonisothermal conditions: nonisothermal consolidation through an unsaturated elastic stratum. Here, we aim to characterize the drained bulk modulus $K_{dr}$, the permeability $k$, and the thermal conductivity $\lambda_{\text{avg}}$, under nonisothermal two-phase fluid flow. Many studies, including \citet{14-THM-1981}, \citet{8-THM-1996}, \citet{13-THM}, and recently \citet{khoei2021modeling} solved the forward problem under various conditions, including isothermal/nonisothermal, swelling/consolidation, and considering/ignoring the effect of phase change. We also recently solved the forward problem using PINNs \cite{amini2022physics}, and here we address the inverse problem for the first time.

The problem setup is based on \citet{khoei2021modeling}, as follows. A ${{10}~\text{cm}}$ stratum with initial conditions of ${{p_c=280}~\text{kPa}}$, ${{p_g=102}~\text{kPa}}$, ${{T=10}~{^\circ}\text{C}}$ is subjected to ${{15}~{^\circ}\text{C}}$ and ${{140}~\text{kPa}}$ increase in temperature and capillary pressure, respectively. The boundary conditions of the top surface are ${{p_c=420}~\text{kPa}}$, ${{p_g=102}~\text{kPa}}$, and ${{T=25}~{^\circ}\text{C}}$. The bottom surface is impermeable to heat transfer and fluid flow, and has fixed displacements.

Material properties include the Young modulus ${E=60~\text{MPa}}$, the Poisson ratio ${{\nu}={0.2857}}$, the porosity ${\phi=0.5}$, and Biot modulus ${b=1}$. The bulk moduli are considered as ${K_s}=0.14 \times 10^{10}~\text{Pa}$, ${K_w}=0.43 \times 10^{13}~\text{Pa}$, and $K_g=0.1 \times 10^{6}~\text{Pa}$ for solid, water, and gas phases, respectively. The fluid flow parameters are taken as $k={6}{\times}{10^{-15}~{\text{m}^2}}$ and ${\mu_w}={\mu_g}={10^{-3}}~{\text{Pa}\cdot \text{s}}$. The densities are ${\rho_s}={1800}~{\text{kg}/\text{m}^3}$, ${\rho_w}={1000}~{\text{kg}/\text{m}^3}$, and ${\rho_g}={1.22}~{\text{kg}/\text{m}^3}$. The thermal conductivity coefficient is considered as ${\lambda_\text{avg}}={0.458}~{\text{J}/\text{s} \cdot \text{m} \cdot {^\circ}\text{C}}$, with the heat capacity as ${C_s}={125460}~{\text{J}/\text{kg} \cdot {^\circ}\text{C}}$, ${C_w}={4182}~{\text{J}/\text{kg} \cdot {^\circ}\text{C}}$, and ${C_g}={1000}~{\text{J}/\text{kg} \cdot {^\circ}\text{C}}$. In addition, the thermal expansion coefficients are given as ${\beta_s}={9}{\times}{10^{-7}}~{1/{^\circ}\text{C}}$, ${\beta_w}={6.3}{\times}{10^{-6}}~{1/{^\circ}\text{C}}$, and ${\beta_g}={3.3}{\times}{10^{-3}}~{1/{^\circ}\text{C}}$. 
To describe the flow of two immiscible fluids through the porous medium, the Brooks--Corey \cite{Brooks-Corey} relations for capillary pressure and relative permeability of the fluid phases are used, as:
\begin{align}
    {p_c}={p_b}{S_e}^{-1/\lambda}, \quad {k_{rw}}={S_e}^{(2+3\lambda)/\lambda}, \quad 
    {k_{rg}}={(1-{S_e})^2}{(1-{S_e}^{(2+\lambda)/\lambda})}, \quad  {S_e}={\frac{S_w - S_{rw}}{1-{S_{rw}}}}, 
\end{align}
in which ${S_e}$ denotes the effective saturation, and ${k_{rw}}$ and ${k_{rg}}$ are the relative permeabilities of water and gas phases, respectively. Finally, the pore size distribution and the residual water saturation parameters are ${\lambda=2.308}$ and ${S_{rw}=0.3216}$, respectively, and the capillary entry pressure is taken as ${{p_b=133.813}~\text{kPa}}$.

\begin{figure}[H]
    \centering
    \includegraphics[width=0.6\textwidth]{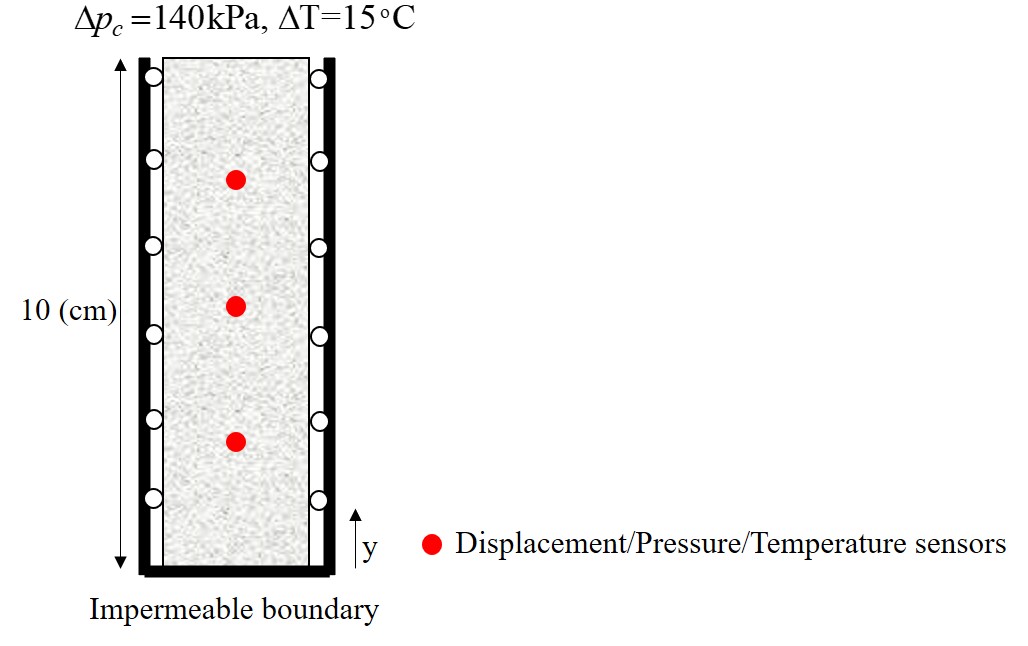}
    \caption{Thermoelastic consolidation of an unsaturated stratum; geometry, boundary conditions and sensor locations are highlighted in the figure.} 
    \label{figs:THM_geometry}
\end{figure}

Given the problem setup and material properties, we take the forward FEM solution for capillary pressure, displacement, and temperature from \citet{khoei2021modeling}, at the first, second, and third quarter of the stratum height, as shown in \cref{figs:THM_geometry} as the ``sensor'' data. Given this data, we then attempt to identify the permeability, drained bulk modulus, and thermal conductivity of the stratum using the proposed inverse PINN--THM solver. 
With the mentioned material properties, the gas pressure variation is about 20~Pa, which is negligible in comparison to capillary and water pressure. Hence, we turn the general format of the equations into a form compatible for passive gas pressure in two-phase fluid flow systems \cite{mohammadnejad2013hydro, schrefler2002mechanics}. 
Therefore, given the constant gas pressure assumption, we can remove the gas fluid flow equation and only consider the capillary pressure as the main functional for the fluid flow, so the water pressure is defined based on the relation \cref{pc_relation}.
Henceforth, the final form of the formulation used for this problem is expressed as:
\begin{align}
\begin{split}
    & {\bar{K}_{dr}}\frac{\partial\bar{\varepsilon}_v}{\partial {\bar{y}}}+{2}\left({\frac{1-2\nu}{1+\nu}}\right){\bar{K}_{dr}}\frac{\partial^2 \bar{u}}{\partial \bar{y}^2} - {b}{\sum_{\alpha}}\frac{\partial}{\partial \bar{y}}{({S_{\alpha}}{\bar{p}_{\alpha}})} - {\bar{\beta}_s}{\bar{K}_{dr}}{N_T}\frac{\partial \bar{T}}{\partial \bar{y}}  = {0}, \label{eq44}
\end{split} \\
\begin{split}
    & -\frac{\bar{\mu}_{w}}{\bar{k}}{({\bar{N}_{ww }}+\frac{b_{w} b_w}{\bar{K}_{dr}}){D^*}{\frac{\partial \bar{p}_c}{\partial \bar{t}}}} + \frac{\bar{\mu}_{w}}{\bar{k}}{\frac{b S_{w}}{\bar{K}_{dr}}}{D^*} \frac{\partial \bar{\sigma}_v}{\partial \bar{t}} - \frac{\bar{\mu}_{w}}{\bar{k}}{\phi S_{w}}\left({\bar{\beta}_{w}}-{\bar{\beta}_s}\right){Q^*} \frac{\partial \bar{T}}{\partial \bar{t}} \\ 
& ~~~~~~~~~~~~~~~~~~ - \frac{\partial}{\partial \bar{y}}{{\left({k_{w}^r}\frac{\partial \bar{p}_w}{\partial \bar{y}}\right)}} = {{0}},  \label{eq45}
\end{split}\\
& {H^*}{\frac{\partial \bar{T}}{\partial \bar{t}}} - {J^*}{\left(
{\bar{\rho}_w}{\bar{C}_w}{k^r_{w}}\frac{\bar{k}}{\bar{\mu}_w} \frac{\partial \bar{p}_w}{\partial \bar{y}} \right)} \frac{\partial \bar{T}}{\partial \bar{y}}- {F^*}\frac{\partial}{\partial \bar{y}}{({{\bar{\lambda}_\text{avg}}}\frac{\partial\bar{T}}{\partial \bar{y}})} = {0}. \label{eq46}
\end{align}
By experimentation, we also realized that for characterization of permeability in this two-phase flow example, multiplying the fluid flow governing equation by ${\bar{\mu}_{w}}/{\bar{k}}$ results in more accurate parameters, i.e., \cref{eq45}. 
We associated this to the small diffusion term in the flow relation as a result of small value for ${k^r_w}$.
Regarding the neural network hyperparameters, the displacement, pressure, and temperature networks are all built with 40 neurons width and with 8 hidden layers and tangent-hyperbolic activation function. A structured sampling grid of 41$\times$100 is used to train the problem. 

\begin{figure}[H]
    \centering
    \includegraphics[width=0.8\textwidth]{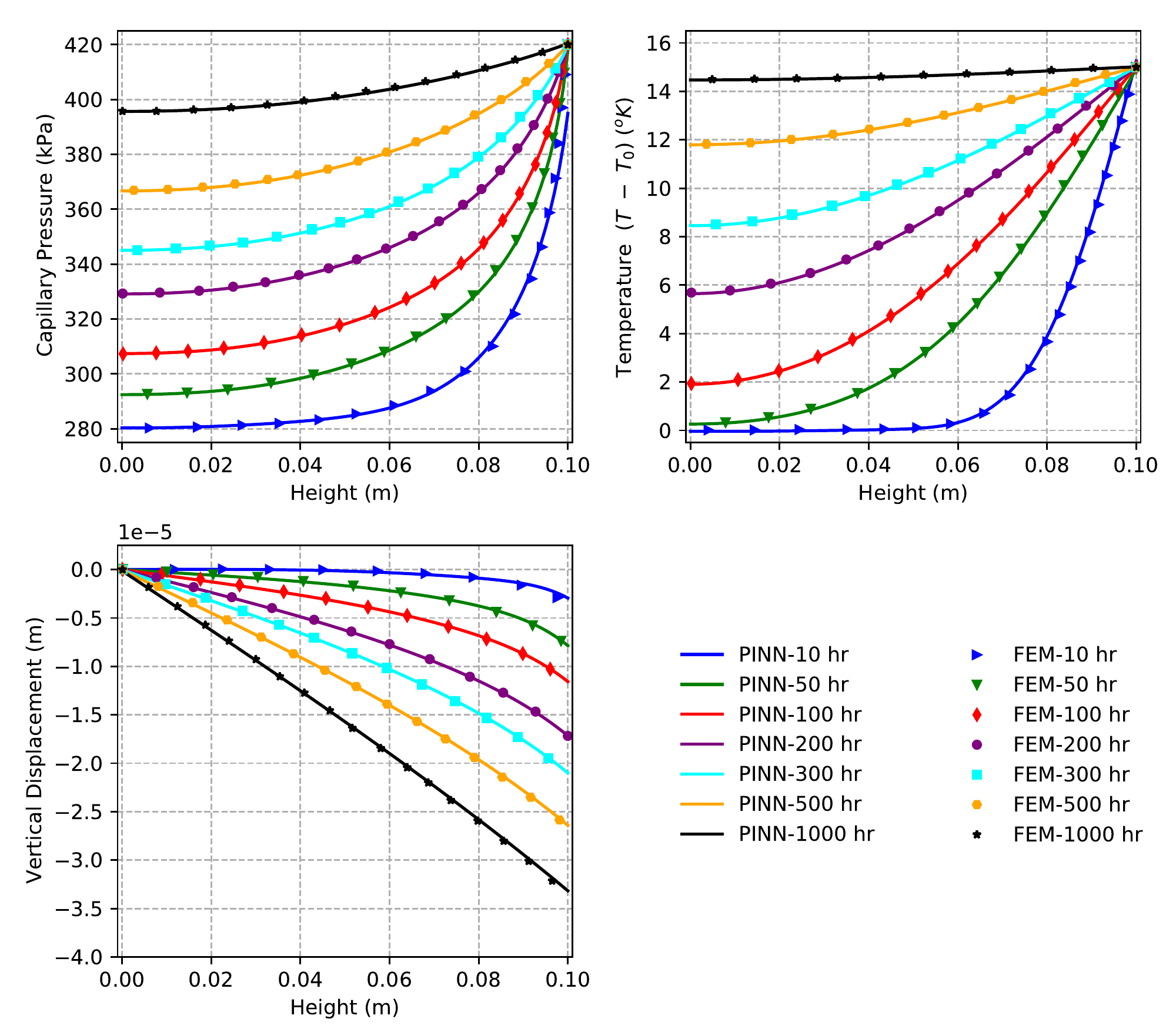}
    \caption{Thermoelastic consolidation of an unsaturated stratum; PINN solutions, i.e., capillary pressure, temperature, and vertical displacement (solid lines) and comparison against the FEM solutions reported by \citet{khoei2021modeling} (symbols). The horizontal axis indicates the depth of the stratum. Different colors show the solution at different times.}
    \label{figs:THM_solution}
\end{figure}

\begin{figure}[H]
    \centering
    \includegraphics[width=1\textwidth]{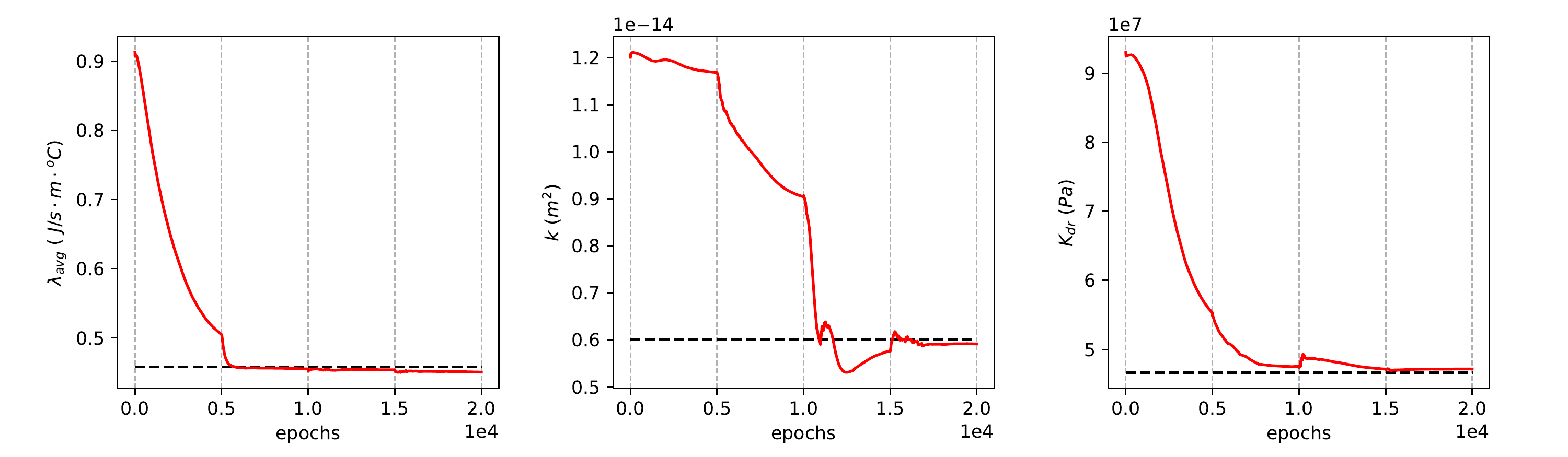}
    \caption{Thermoelastic consolidation of an unsaturated stratum; the evolution of characterization of average thermal conductivity $\lambda_{\text{avg}}$, permeability ${k}$, and bulk compressibility ${K_{dr}}$parameters evolution as a function of training epochs, with vertical lines highlighting the sequential iterations.}
    \label{figs:THM_parameters}
\end{figure}

The results of the sequential PINN--THM inverse solver are shown in \cref{figs:THM_solution,figs:THM_parameters}. With initial values far from their true values, we find that the proposed solver results in highly accurate temperature, displacement, and capillary pressure spatio-temporal distributions compared with FEM results, as shown in \cref{figs:THM_solution}. The evolution of the estimated parameter values, shown in \cref{figs:THM_parameters}, additionally highlights the success of sequential training for this complex inverse problem. 
Compared to the previous problems, this example exhibited the highest sensitivity with respect to the medium fluid-flow permeability and to the choices of optimization hyperparameters such as learning rate---a finding that is not entirely surprising given the complexity and nonlinearity of the coupled PDEs. 

\section{Conclusions}
We investigated the application of sequential PINN--THM for the characterization of deforming fully and partially saturated porous media under isothermal and nonisothermal conditions. 
The proposed PINN framework features a dimensionless form of the THM governing PDEs suitable for inverse analysis, a sequential training strategy for the multi-physics THM processes in porous media, and adaptive weight methods to prevent unbalance convergence of loss terms. We validated our proposed framework on three reference problems of porous media, and reported remarkable performance of PINNs to simultaneously capture the solution variables and identify the PDE parameters. 

Model inversion using classical methods is a nontrivial and computationally expensive task. Classical solvers such as FEM and FVM are well suited to simulate forward problems but an external optimization loop is needed to perform inversion. Each iteration of the optimization requires loss and gradient evaluation and, therefore, multiple forward solves are needed per optimization iteration to perform a parameter update. The inversion optimization can also suffer from the complex loss landscape associated with multidimensional, coupled and highly nonlinear PDEs. In this respect, the main advantage of PINN solvers is that they can be used for simultaneous forward and inverse modeling, as data incorporation, loss and gradient evaluations are naturally a part of PINN solvers. Therefore, while in general the performance of PINN forward solvers is far from optimal, we find that as inverse solvers they can in fact be competitive. Substantial performance improvements, however, are likely as PINN network architectures and optimization strategies continue to evolve. 

The applications of our proposed PINN--THM inversion framework have so far been limited to synthetic problems on simple geometries. In the future, we plan to apply the framework to more complex problems with real-world, multifaceted datasets.

\section*{Data availability statement}
All data, models, or code generated or used during the study are available in a repository online (\href{https://github.com/sciann/sciann-applications/tree/master/SciANN-PoroElasticity}{https://github.com/sciann/sciann-applications/tree/master/SciANN-PoroElasticity}) in accordance with funder data retention policies.

\section*{Acknowledgements}
RJ was partly funded by the KFUPM-MIT collaborative agreement `Multiscale Reservoir Science'.



\bibliographystyle{plainnat}

\bibliography{references}

\end{document}